\documentclass[conference]{IEEEtran}

\IEEEoverridecommandlockouts
\usepackage[numbers, sort]{natbib}
\usepackage{amsmath,amssymb,amsfonts}
\usepackage{algorithmic}
\usepackage{graphicx}
\usepackage{subcaption}
\usepackage{textcomp}
\usepackage{xcolor}
\usepackage{caption}
\usepackage{siunitx}
\usepackage{url}
\usepackage{hyperref}
\hypersetup{
		pdftitle={Skin Lesions Classification Using Convolutional Neural Networks in Clinical Images}, 
		pdfauthor={D. B. Mendes; N. C. Silva},
    	pdfsubject={Skin Lesion},
	    pdfcreator={D. B. Mendes},
		pdfkeywords={Deep Learning, Skin Lesion, ResNet-152, Dermatology}, 
		colorlinks=true,       		% false: boxed links; true: colored links
    	linkcolor=black,          	% color of internal links
    	citecolor=black,        		% color of links to bibliography
    	filecolor=black,      		% color of file links
		urlcolor=black,
		bookmarksdepth=4
}
\def\BibTeX{{\rm B\kern-.05em{\sc i\kern-.025em b}\kern-.08em
    T\kern-.1667em\lower.7ex\hbox{E}\kern-.125emX}}
    
\newcommand{\source}[1]{\caption*{Source: {#1}} }
\newcommand{\citetextd}[1]{\citeauthor{#1} (\citeyear{#1})}

\begin{document}

\makeatletter
\title{Skin Lesions Classification Using Convolutional Neural Networks in Clinical Images%\\
}\let\Title\@title
\author{\IEEEauthorblockN{Danilo Barros Mendes *}
\IEEEauthorblockA{\textit{Faculdade de Engenharias do Gama, FGA} \\
\textit{University of Brasília, UnB}\\
Gama, Brazil \\
dan.b412@gmail.com}
\and
\IEEEauthorblockN{Nilton Correia da Silva *}
\IEEEauthorblockA{\textit{Faculdade de Engenharias do Gama, FGA} \\
\textit{University of Brasília, UnB}\\
Gama, Brazil \\
niltoncs@unb.br}
% \and
% \IEEEauthorblockN{Teófilo Emidio de Campos *}
% \IEEEauthorblockA{\textit{Campus Darcy Ribeiro} \\
% \textit{University of Brasília, UnB}\\
% Brasília, Brazil \\
% teodecampos@unb.br}
\thanks{* The author is a member of the ``Machine Learning Research Group'' from the University of Brasília. Access \url{www.gpam.unb.br} for more information.}
% \and
% \IEEEauthorblockN{Given Name Surname}
% \and
% \IEEEauthorblockN{Given Name Surname}
% \and
% \IEEEauthorblockN{Given Name Surname}
}
\makeatother

\maketitle

\begin{abstract}
Skin lesions are conditions that appear on a patient due to many different reasons. One of these can be because of an abnormal growth in skin tissue, defined as cancer. This disease plagues more than 14.1 million patients and had been the cause of more than 8.2 million deaths, worldwide. Therefore, the construction of a classification model for 12 lesions, including Malignant Melanoma and Basal Cell Carcinoma, is proposed. Furthermore, in this work, it is used a ResNet-152 architecture, which was trained over 3,797 images, later augmented by a factor of 29 times, using positional, scale, and lighting transformations. Finally, the network was tested with 956 images and achieve an area under the curve (AUC) of 0.96 for Melanoma and 0.91 for Basal Cell Carcinoma.
\end{abstract}


\section[Introduction]{Introduction}

Today, skin cancer is a public health and economic issue, that for long years have been approached with the same methodology by the dermatology field \cite{hamblin2016imaging}. This is troublesome when we analyze that for the last 30 years the numbers of cases diagnosed with skin cancer have increased significantly \cite{americansociety2018}. It is more troublesome when money comes in the equation, seeing that millions of dollars are being spent in the public sector \cite{DeSouza2011}. A major part of this is spent in the individual analysis of the patient. Where the doctor analyzes the lesion and takes action on the pieces of evidence seen. If any of these steps were to be optimized, it could mean a decrease in expenditure for the whole dermatology sector. 

Dermatology is one of the most important fields of medicine, with the cases of skin diseases outpacing hypertension, obesity and cancer summed together. That is accounted because skin diseases are one of the most common human illness, affecting every age, gender and pervading many cultures, summing up to between 30\% and 70\% of people in the United States. This means that in any given time at least 1 person, out of 3, will have a skin disease \cite{bickers2006burden}. Therefore, skin diseases are an issue on a global scale, positioning on 18th in a global rank of health burden worldwide \cite{hay2014global}.

Furthermore, medical imaging can show itself as a resource of high value, as dermatology has an extensive list of illness that it has to treat. In addition, the field has developed its own vocabulary to describe these lesions. However, verbal descriptions have their limitations and a good picture can replace successfully many sentences of description and is not susceptible to the bias of the message carrier.

In addition, the recommended way to detect early skin diseases is to be aware of new or changing skin growths \cite{americansociety2017}. Analysis with the naked eye is still the first resource used by specialists, along with techniques such as ABCDE, that consists of scanning the skin area of interest for asymmetry, border irregularity, uniform colors, large diameters and evolving patches of skin over time \cite{nachbar1994abcd}. In this way, the analysis from medical images is analogous to the analysis with the naked eye and thus can be applied the same techniques and implications. This supports the idea that skin cancer often is detectable through naked eye and medical photography.
% However, these techniques are usually not used to diagnose a patient with an illness or not. 

Worldwide the most common case of cancer is skin cancer, been melanoma, basal and squamous cell carcinoma (BCC and SCC) the most frequent types of the disease \cite{americansocietybcc}. This type of the disease is most frequent in countries with the population with predominant white skin or in countries like Australia or New Zealand \cite{stewart2014world}. 

In Brazil, it is estimated that for the biennium of 2018-2019, there will be \num{165580} new cases of non-melanoma skin cancer (BCC and SCC mostly) \cite{sensoinca2018}. Moreover, it is visible that the incidence of these types of skin cancer had risen for many years. This increase can be due to the combination of various factors, such as longer longevity of the population, more people being exposed to the sun and better cancer detection \cite{americansocietybcc}.

In the United States, the numbers add up to \num{9730} deaths estimated for 2017 \cite{americansociety2017}. Skin cancer accounts for more than \num{1688780} cases (not including carcinoma in situ, nor non-melanoma cancers) in the US alone in the year of 2017 \cite{americansociety2017}.

Despite skin cancer being the most common type of cancer in society, it does not represent a great death rate in its first stages, since the patient has a survival rate of $97\%$. However, if the patients are diagnosed in the later stages the 5-year survival rate decreases to $15\%$.

In Brazil, were expected to occur \num{114000} new cases of non-melanoma skin cancer in 2010. From that, it was expected that $95\%$ were diagnosed in early stages. However, even with early diagnosis this amount of cases means around R\$$37$ million (Reais) to the health public system and R\$$26$ million to the private system per year \cite{DeSouza2011}.

\begin{figure*}
    \includegraphics[width=\textwidth]{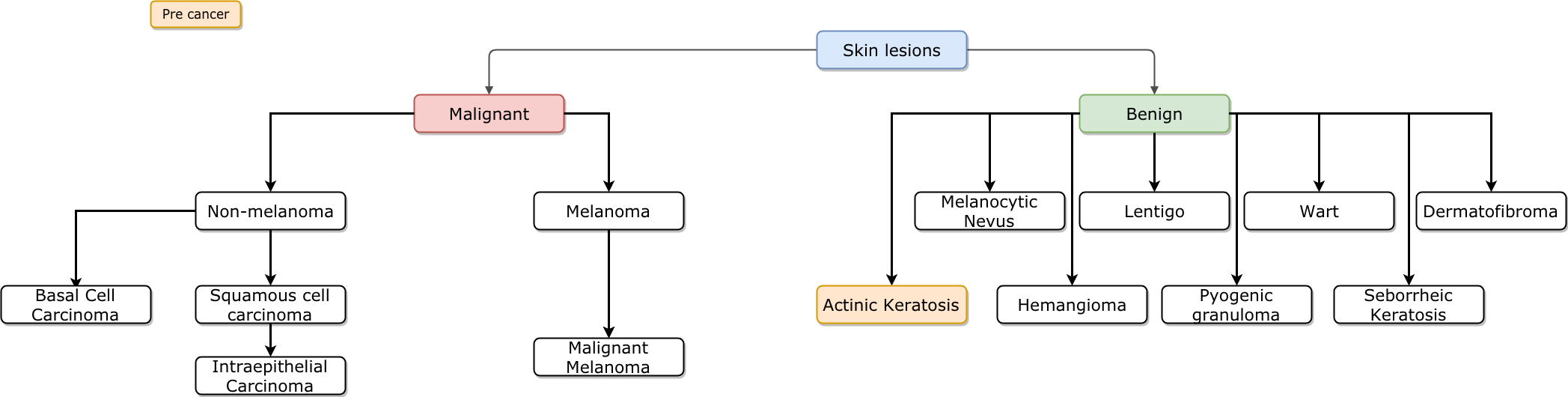}
    \caption{Skin lesions groups.}
    \source{Authors.}
    \label{fig:lesiondiagram}
\end{figure*}

Moreover, we can divide skin lesions into two major groups, one being malignant lesions and the other benign lesions. The first is composed mostly of skin cancers and the latter being composed of any lesion that does not pose a major threat. One counterexample of this division is the actinic keratosis, that presents itself as a potential SCC, as it has the potential to develop into it. Thus, actinic keratosis is classified as a precancerous lesion \cite{Prajapati699}. Furthermore, this work analyzed and chose \num{12} lesions in total, 4 malignant and 8 benign (being 1 precancerous), as seen on Figure \ref{fig:lesiondiagram}. The lesions where chosen mainly on the public data available online, to be described in subsection \ref{sec:datasets}.

% Artificial intelligence has the potential to analyze a lot of images and perform difficult classifications on it, helping the diagnosis of certain illness. Furthermore, detecting skin lesions are mainly done by scanning the patient with the naked eye and then execute different approaches to finally diagnose the patient. This expresses a major task of classification as the specialist tries to fit the lesion in a broad spectrum of possibilities, given only the symptoms and the appearance of the lesion in the skin.

Seeing the problems involved in diagnosing skin lesions, this work envisions to create a learning model to classify skin lesions in one of 12 conditions of interest. With this purpose, the classifier aims to correct distinguish lesions analyzing clinical images with the condition. Furthermore, this can prove to be a useful tool to aid patients and doctors on a daily basis operation.

Furthermore, this work is done with the vision of being the stepping stone for newer approaches that democratize and distribute access to health care. A good lesion classification model may be the motor that will accelerate the construction of tools that puts the possibility of early diagnosis and alert on patient`s hands, even far isolated patients, where few doctors can reach. These tools may save many lives and reduce several costs with the treatment of late-stage diseases.

The related work in this field proved that there are many algorithms capable of tackling this problem, but there is an astonishing difference between shallow and deep methods in machine learning. With that in view, this work will guide its efforts in using deep neural networks to achieve its main objective. For this to happen, the gathering of good practices and techniques used to approach the classification of clinical images is needed.
\section{Related Work} \label{sec:related_work}

% The problems in detecting skin lesions accurately are that it exists many features and minutiae that need to be dealt with. For that task, many professionals train for part of their life to specialize in detecting and differentiating these diseases. Is not uncommon that one may dedicate a lifelong effort to continually improve one's ability. However specialized one may be, one is still susceptible to human failures such as fatigue and mistakes. Another limitation that a specialist faces is the workload that is possible to take in any given day, for such is human nature to not be able to work several hours a day and process many pieces of information at a fast rate, not staggering once. 

Many algorithms and tools have been created to aid these professionals in their task of detecting diseases in many fields \cite{aerts2014decoding,Esteva2017,lee2017fully,vandenberghe2017relevance,Kermany2018}. This has proven to add more reliability and confidence to doctors in their practices as they have more information to diagnose patients. For dermatology and skin lesions detection has not been different. History shows that many approaches had been made over the course of years, applications with shallow algorithms such as K-Nearest Neighbors (KNN) \cite{ballerini2013color} and Support Vector Machines (SVM) \cite{gilmore2010support} had been proven to accomplish good results, but are as well tiresome to build applications that involve such approaches.

Seeing this, some researchers have been applying this approach to classifying skin lesions with success. One common thing in this domain is the lack of quality and scarcity of open data. It is common to see works with only a couple hundred of examples. That is a characteristic of the medical field. There are many hospitals and clinics that hold huge amounts of data and do not make it public mainly because of privacy issues with patients. However, many authors still apply efforts to push forward the technology in such fields, overcoming these barriers. For the purposes of this work, we listed some related researches that uses deep learning in dermatology, applying neural networks to skin lesions.

\citetextd{Matsunaga} proposed an approach to classify melanoma, seborrheic keratosis, and nevocellular nevus, using dermoscopic images. In their work, they proposed an ensemble solution with two binary classifiers, that still leveraged from age and sex information of the patients, if they were available. Furthermore, they utilized techniques of data augmentation, using a combination of 4 transformations (rotation, translation, scaling and flipping). For the architecture, they chose the ResNet-50 implementation on the framework Keras, with personal modifications. This model was pre-trained with the weights for a generic object recognition model and finally used two optimizers AdaGrad and RMSProp. This work was then submitted to the ISBI Challenge 2017 and won first place, ahead of other 22 competitors.

\citetextd{Nasr-Esfahani2016} showed a technique that uses imaging processing as a previous step before training. This result in a normalization and noise reduction on the dataset, since non-dermoscopic images are prone to have non-homogeneous lightning and thus present noise. Moreover, this work utilizes a pre-processing step using \textit{k-means} algorithm to identify the borders of a lesion and extract a binary mask, which the lesion is present. This is done to minimize the interference of the healthy skin in the classification. Furthermore, \citetextd{Nasr-Esfahani2016} used a technique called data augmentation to increase the dataset, using three transformations (cropping, scaling and rotation) and multiplied the dataset by a factor of 36 times. Finally, a pre-trained convolutional neural network (CNN) is used to classify between melanoma and melanocytic nevus for \num{200} epochs (\num{20000} iterations, using a batch size of 64 and a dataset with \num{6120} examples). 

\citetextd{Menegola2017} presented a thorough study for the 2017 ISIC Challenge in skin-lesion classification. In this work, it is presented experimentations with some pre-trained deep-learning models on ImageNet for a three-class model classifying melanoma, seborrheic keratosis, and other lesions. Models such as ResNet-101 and Inception-v4 were vastly experimented with several configurations of the dataset, utilizing 6 data sources for the composition of the final dataset. It was also reported the use of data-augmentation with at least 3 different transformations (cropping, flipping, and zooming). Also, it is reported that the points that were critical to the success of the project were mainly due to the volume of data gathered, normalization of the input images and utilizing meta-learning. The latter is elucidated as an SVM layer in the final output of the deep-learning models, that map the outputs to the three classes that were proposed in the challenge. Finally, this work won the first place in the 2017 ISIC Challenge for skin lesion classification.

\citetextd{Kwasigroch} present a solution similar to the previous 3. This is due to the inherent limits and problems that are existent in this domain, data scarcity. In this work transfer-learning is applied, using two different learning models, VGG-19 and ResNet-50, both pre-trained on ImageNet \num{1000} classes dataset. These were used to classify between malignant and benign lesions, using \num{10000} dermoscopic images. For the correct learning process, it was also used the up-sampling of the underrepresented class. This process was done using a random number of transformations, chosen between rotation, shifting, zooming, and flipping. Furthermore, in this paper, it was presented 3 experiments, first with the VGG-19 architecture with the addition of two extra convolutional layers, two fully connected layers, and one neuron with a sigmoid function. Second it experimented with the ResNet-50 model, and finally a implementation of VGG-19 with an SVM classifier as the fully-connected layer. As a final result, the modified implementation of the VGG-19 had the best results. However, the main reason for the poor results in the ResNet-50 model was due to the small amount of training data. Maybe with larger amounts of data, it would be possible to train a small model and produce better results.

\citetextd{Esteva2017} presented a major breakthrough in the classification of skin lesions. This research compared the result of the learning model with 21 board-certified dermatologists and proven to be more accurate in this task. It was performed to classify clinical images, indicating whether a lesion is a benign or malignant one. For this result were used \num{129450} images, consisting of \num{2032} different diseases and including \num{3372} dermoscopic images. Furthermore, it was used a data-augmentation approach to mitigate problems as variability in zoom, angle, and lighting present in the context of clinical images. The augmentation factor was by 720 times, using rotation, cropping, and flipping. Here, an Inception-v3 pre-trained model was utilized as the main classifier, fine-tuning every layer and training the final fully connected layer. Moreover, the training was done for over than 30 epochs using a learning rate of \num{0.001}, with a decay of 16 after every 30 epochs. The classification was done in such a way that the model was trained to classify between 757 fine-grained classes, and then as the probabilities were predicted it was fed into an algorithm that selected the two different classes (malignant or benign). Using this approach, this work achieved a new state of the art result.

\citetextd{SeogHan2018} proposed to classify the skin lesions as unique classes, not composing meta-classes such as benign and malignant. It used the ResNet-152 pre-trained on the ImageNet model to classify 12 lesions. However, for training was used other 248 additional classes, that were added to decrease the false positive and improve the analysis of the middle layers of the model. Furthermore, this was done in such a way that the train sampling for the 248 diseases did not outgrow the main 12, thus when used for inference the model predicted one of the 12 illness, even when the lesion does not belong to one of them. For training was used \num{855370} images, augmented approximately 20 to 40 times, using zooming and rotation. These images were gathered from two Korean hospitals, two publicly available and biopsy-proven datasets, and one dataset constructed from 8 dermatologic atlas websites. Furthermore, the training lasted for 2 epochs using a batch size of 6 and a learning rate of \num{0.0001} without decay before 2 epochs. This early stopping was done to avoid overfitting on the dataset. Finally, it was reported that the ethnic differences presented in the context were responsible for poor results in different datasets, thus it was necessary to gather data from different ethnics and ages to correct mold the solution to reflect the real world problem present in skin lesions classification.

Finally, we can observe that every one of these works has one aspect in common, data scarcity. This is a characteristic of the medical domain, there are very few annotated examples of data that are publicly available. The works that proven to have more impact had to collect data from other sources, mainly private hospitals or clinics. Furthermore, this step of data collection did not fully mitigate the problem, it was still necessary to use techniques such as transfer-learning \cite{pan2010survey,yosinskiCBLtransfer2014} and data-augmentation \cite{simard2003best,van2001art,krizhevsky2012imagenet}. 
\section[Methods and Materials]{Methods and Materials}\label{ch:skin}

% Seeing the past history of dermatology and the implementation of artificial intelligence in the field, it is safe to say that many characteristics have evolved and developed over the years. It is also correct to state that the current approach has had the most promising results, but it has a long way to go until it is viable to use inside clinics and at homes. For this purpose, it is necessary to understand the underlining concepts underneath the current state of the art. 

% Furthermore, this chapter will discuss these concepts in a succinct manner, going through the raw material needed to construct it, knowledge about the technology that is implemented in the field of machine learning, how to tie the technology and the data together in a medical field, such as dermatology, what is the current state of the industry of artificial intelligence, and finally, how to integrate everything in one piece.

\subsection{Datasets}\label{sec:datasets}

Due to the scarcity of data present in the medical field, the datasets chosen were not the selection of the best on a collection of options. The process of choosing one mainly took into account the criterion of public availability. Aside from that, the only pre-requisite was that the dataset was composed with only clinical images (photos taken from cameras without other tools or distorting lenses).

From these criteria, only two datasets fitted the description. The datasets contained 10 (ten) distinct lesions, containing 4 malignant illnesses at maximum. Another additional dataset was gathered from dermatologic websites, using a script for scrapping pages. The latter dataset was acquired from the work of \citetextd{SeogHan2018} and is not publicly available due to copyrights owned by the websites. Finally, these datasets are further discussed below.

\subsubsection*{MED-NODE}
The first dataset used is provided by the Department of Dermatology at the University Medical Center Groningen (UMCG) \cite{giotis2015med}. This dataset contains 170 images that are divided between 70 melanoma and 100 nevus cases. Furthermore, these images were processed with an algorithm for hair removal.

\subsubsection*{Edinburgh}
The second dataset is provided by the Edinburgh Dermofit Image Library \cite{ballerini2013color} and is publicly available for purchase, under an agreement with the license of use\footnote{Available at \url{https://licensing.eri.ed.ac.uk/i/software/dermofit-image-library.html}.}. This dataset is the more complete one found on the web. It contains \num{1300} images, that are divided into 10 lesions, including melanoma, BCC, and SCC. These images are all diagnosed based on experts opinions. In addition, it is also provided the binary segmentation of the lesion, for each one. It is valid to note that the images are not all in the same size.

Furthermore, the lesions and its respective numbers are listed in the table \ref{tab:lesions_edin}. 

\begin{table}[h]
 \centering
% distancia entre a linha e o texto
 {\renewcommand\arraystretch{1.25}
 \caption{Lesion sampling for Edinburgh dataset.}
 \label{tab:lesions_edin}
 \begin{tabular}{ l l }
  \cline{1-1}\cline{2-2}  
    \multicolumn{1}{|c|}{\textit{\textbf{Lesion Type}}} &
    \multicolumn{1}{c|}{\textit{\textbf{Number of images}}}
  \\  
  \cline{1-1}\cline{2-2}  
    \multicolumn{1}{|c|}{\textit{Actinic Keratosis}} &
    \multicolumn{1}{c|}{\textit{45}}
  \\  
  \cline{1-1}\cline{2-2}  
    \multicolumn{1}{|c|}{\textit{Basal Cell Carcinoma}} &
    \multicolumn{1}{c|}{\textit{239}}
  \\  
  \cline{1-1}\cline{2-2}  
    \multicolumn{1}{|c|}{\textit{Melanocytic Nevus (mole)}} &
    \multicolumn{1}{c|}{\textit{331}}
  \\  
  \cline{1-1}\cline{2-2}  
    \multicolumn{1}{|c|}{\textit{Seborrhoeic Keratosis}} &
    \multicolumn{1}{c|}{\textit{257}}
  \\  
  \cline{1-1}\cline{2-2}  
    \multicolumn{1}{|c|}{\textit{Squamous Cell Carcinoma}} &
    \multicolumn{1}{c|}{\textit{88}}
  \\  
  \cline{1-1}\cline{2-2}  
    \multicolumn{1}{|c|}{\textit{Intraepithelial Carcinoma}} &
    \multicolumn{1}{c|}{\textit{78}}
  \\  
  \cline{1-1}\cline{2-2}  
    \multicolumn{1}{|c|}{\textit{Pyogenic Granuloma}} &
    \multicolumn{1}{c|}{\textit{24}}
  \\  
  \cline{1-1}\cline{2-2}  
    \multicolumn{1}{|c|}{\textit{Haemangioma}} &
    \multicolumn{1}{c|}{\textit{97}}
  \\  
  \cline{1-1}\cline{2-2}  
    \multicolumn{1}{|c|}{\textit{Dermatofibroma}} &
    \multicolumn{1}{c|}{\textit{65}}
  \\  
  \cline{1-1}\cline{2-2}  
    \multicolumn{1}{|c|}{\textit{Malignant Melanoma}} &
    \multicolumn{1}{c|}{\textit{76}}
  \\  
  \cline{1-1}\cline{2-2}  
    \multicolumn{1}{|c|}{\textbf{\textit{TOTAL}}} &
    \multicolumn{1}{c|}{\textbf{\textit{1,300}}}
  \\  
  \hline
 \end{tabular} }
\end{table}

\subsubsection*{Atlas}

This last dataset, was acquired from running several scripts for scrapping different dermatological websites\footnote{These websites included, \url{http://dermquest.com}, \url{http://www.dermatlas.net}, \url{http://www.dermis.net/dermisroot/en/home/index.htm},
\url{http://www.meddean.luc.edu/lumen/MedEd/medicine/dermatology/ melton/atlas.htm}, \url{http://www.dermatoweb.net}, \url{http://www.danderm-pdv.is.kkh.dk/atlas/index.html}, \url{http://www.atlasdermatologico.com.br}, \url{http://www.hellenicdermatlas.com/en}.}. So that is the reason that this dataset was baptized as Atlas. This dataset was obtained from \citetextd{SeogHan2018} in a personal submitted request. It contains \num{3816} images downloaded from websites and distributed between six lesions.

The difference from the Edinburgh dataset is that this contains two lesions that are not present on the first, Wart and Lentigo -- both benign lesions --, as it can be seen on table \ref{tab:lesions_atlas}. This, alongside with the Atlas and MED-NODE datasets, sums up to 12 lesions, that are the interest of this work. 

\begin{table}[h]
 \centering
% distancia entre a linha e o texto
 {\renewcommand\arraystretch{1.25}
 \caption{Lesion sampling for Atlas dataset.}
 \label{tab:lesions_atlas}
 \begin{tabular}{ l l }
  \cline{1-1}\cline{2-2}  
    \multicolumn{1}{|c|}{\textit{\textbf{Lesion Type}}} &
    \multicolumn{1}{c|}{\textit{\textbf{Number of images}}}
  \\   
  \cline{1-1}\cline{2-2}  
    \multicolumn{1}{|c|}{\textit{Basal Cell Carcinoma}} &
    \multicolumn{1}{c|}{\textit{1,561}}
  \\  
  \cline{1-1}\cline{2-2}  
    \multicolumn{1}{|c|}{\textit{Lentigo}} &
    \multicolumn{1}{c|}{\textit{69}}
  \\  
  \cline{1-1}\cline{2-2}  
    \multicolumn{1}{|c|}{\textit{Malignant Melanoma}} &
    \multicolumn{1}{c|}{\textit{228}}
  \\  
  \cline{1-1}\cline{2-2}  
    \multicolumn{1}{|c|}{\textit{Melanocytic nevus (mole)}} &
    \multicolumn{1}{c|}{\textit{626}}
  \\  
  \cline{1-1}\cline{2-2}  
    \multicolumn{1}{|c|}{\textit{Seborrheic keratosis}} &
    \multicolumn{1}{c|}{\textit{897}}
  \\  
  \cline{1-1}\cline{2-2}  
    \multicolumn{1}{|c|}{\textit{Wart}} &
    \multicolumn{1}{c|}{\textit{435}}
  \\
  \cline{1-1}\cline{2-2}  
    \multicolumn{1}{|c|}{\textbf{\textit{TOTAL}}} &
    \multicolumn{1}{c|}{\textbf{\textit{3,816}}}
  \\  
  \hline
 \end{tabular} }
\end{table}

One difference between Atlas and the first two datasets is the quality of the images, since the dataset was collected from web pages, is not all images that present the same quality, nor the same common viewpoints observed on the Edinburgh dataset. Therefore, this dataset is the most heterogeneous in matters of quality of imaging, viewpoints, the age of patients and ethnicity. However, this dataset in its entirety is not officially diagnosed by specialists, but on the other hand, these photos were displayed on websites that are reliable and used by students. So, there is a heuristic that these images were revised before putting to display in these websites and can be trusted.

\subsection{Handling data scarcity}

As noted previously, for the correct generalization of the weights and biases of a network, a huge amount of data is needed. However, the medical field lacks this amount of images and if only used the data public provided, a good generalization of the problem cannot be met if we wish to train a deep neural network.

% Therefore, the need for new approaches arise. How do we approach a problem that lacks the data needed for the proper training on a blank network? One option is to gather new data. But if it is not possible to do it we ``forge'' new data. But this processes needs to be done in such a way that the final product is not altered to the point that the original label does not fit it anymore. We can call these as non-invasive transformations. That although we alter the original data, the label can still be applied to it. 

% Another option to this is to use previous knowledge of other similar problems and build the new concepts needed on top of it. However, to do this another problem with huge amounts of data is needed. Furthermore, this new problem has to hold some kind of relation to the problem that needs to be solved. This is necessary for the transferability of knowledge itself. For an instance, if we train a network to detect objects, a simple subset of random objects, and then used this built knowledge in a problem to detect faces, this will surely help. It is intuitive to think, if someone knows how to detect an object - a thing - then it surely can be taught to detect a pencil in a table, as the perception to spot edges, curves and differences in colors is already a learned ability. 

% This two options are called \textbf{Data Augmentation} and \textbf{Transfer Learning}, respectively.

\subsubsection{Transfer Learning}

In practice, the domains that are faced in the industry, rather than the academia, usually have low numbers of labeled data. This poses a major obstacle to train a deep convolutional neural network from scratch, since the data may not demonstrate a true representation of the real world. Thus, it is common to see works that utilize the pre-trained weights of a previously trained architecture, this can lead to 2 major approaches. 

The approaches are: using a CNN as a fixed feature extractor or fine-tuning the trained model. The first is mostly used to collect features of images and then use them to train a linear classifier in a new dataset. The second strategy is to continue the training of the network, replacing completely the final layer, but updating the parameters through backpropagation.

A common use of transfer in computer vision, more specifically object classification, is to use pre-trained models that were trained on the ImageNet dataset. Some recent work done by \citetextd{kornblith2018better} shows that ResNets take the lead in performance when treated as feature extractors. While only fine-tuning some models to other datasets, they achieved a new state-of-the-art. All these tests used pre-trained weights and fine-tuned them with Nesterov momentum for \num{19531} steps, which sometimes corresponded as more than \num{1000} epochs using a batch size of \num{256}. Finally, it was proven, empirically, that the Inception-v4 architecture achieves overall better results for this task than the other 12 pre-trained classification models.

Therefore, transfer learning optimizes and cuts short most of the time in the training of new applications. However, this can add some constraints to the work. One example of this is when using a pre-trained network is not possible to extract and change arbitrarily the layers of the network. Another point is that normally, small learning rates are applied to CNN weights that are being fine-tuned. This is because we already expect that the weights are good, and we do not want to distort them too much \cite{yosinskiCBLtransfer2014}.

\subsubsection{Data augmentation}\label{sub:augmentation_transformations}

Data augmentation is a technique used where we do not have an infinite amount of data to train our models. This can be done by introducing random transformations to the data. In image classification, this can be translated as rotating, flipping and cropping the image. These perturbations add more variability to the input, thus this could mean an overfitting reduction in our model by teaching it about invariances in the data domain \cite{krizhevsky2012imagenet, wang2017dataaugmentation, cubuk2018autoaugment}. Therefore, these transformations do not change the meaning of the input, thus, the label originally attributed to it still holds its importance.

Although some transformations in an image can be done agnostic to the field of application (e.g. translation), some other transformations are entitled to domain-specific characteristics. For this work we used an additional transformation that randomizes the natural light effect in the picture, this was done to mimic the transformations seen in indoors clinics due to different light sources. Furthermore, to increase the variability added by augmenting data, the probability of application and magnitude variability are added to the transformations.

Have seen the needs for augmentation, it was used the Augmentor Python library \cite{bloice2017augmentor} for implementing the process of augmenting the dataset. The library has predefined transformations and has a hot-spot for new implementations of transformations. This was quite useful when implementing the method to add light variance to the augmentations.

Each transformation chosen to be applied had been based on general guidelines of data augmentation \cite{wang2017dataaugmentation, cubuk2018autoaugment} or on the nature of the data. Thus, the transformations were aligned in a pipeline fashion, where each had a probability that defined the likelihood of being applied to the image and at the end, the new image was saved in the destination.
% \footnote{The algorithm that implements this pipeline, and the lightning transformation, can be seen in \url{https://github.com/DaniloBarros/SkinLesionClassification}.}
Furthermore, the operations used for this work were the ones listed in table \ref{tab:transformations}.

\begin{table}[ht]
\centering
\caption{Transformations applied for data augmentation.}
\label{tab:transformations}
\begin{tabular}{lr}
\multicolumn{1}{c}{\textbf{Transformation}} & \textbf{Probability} \\ \hline
Rotation                                    & 0.5                  \\
Random zoom                                 & 0.4                  \\
Flip horizontally                           & 0.7                  \\
Flip vertically                             & 0.5                  \\
Random distortion                           & 0.8                  \\
Lightning variance                          & 0.5                 
\end{tabular}
\end{table}

% Seeing this, the algorithm that implements this pipeline can be seen in Appendix \ref{apx:data_augmentation}.

% --------------------------------

% After knowing the possible architectures to use and what are the characteristics of the datasets it was time to gather all this knowledge and implement some solutions. For that, an approach of experimentation was used and different configurations of the dataset and hyperparameters were applied, creating different experiments. However, some parts of this experimentation were similar, such as the preparation of the datasets, metrics used and the infrastructure used to perform these tests. Furthermore, in this work, it is listed only as the most prominent experiment, so that it does not become extremely long.
% For the other experiments, please refer to the Appendixes.

\subsection{Datasets Preparation}

The first thing done, before applying transformations to the dataset, was to separate a test set, usually a 10\% to 20\% of each lesion, depending on the experiment. Following this, if needed for the experiment, was done the data augmentation process. Then the remaining sample was analyzed to see how much was necessary to augment each class. The process augmented the remaining dataset, usually, by a factor of 29 times. That summed with the original dataset was accounted to 30 times the original amount.

After processing the images necessary to compose the training and test datasets, the images for the training dataset were processed to create an LMDB file \cite{chu2011mdb} for fast access to the data in training time. In this process the training dataset is divided between a training set and a validation set. Thus, this split is done in a way that 80\% of the data is used for training and 20\% is for validation. However, this split is done in a stratified way, so that each split has a fair amount of each class.

Finally, these slices of the dataset are kept separated and are used as such for the experiment. % The code that implements the creation of the LMDB file and the split of the training/validation set can be seen in Appendix \ref{apx:create_lmdb}.

\subsection{Architecture}

The architecture used for this work has been the ResNet-152, used with pre-trained weights trained on ImageNet database. This architecture was chosen mainly for the results the family (ResNet-50, ResNet-101, and ResNet-152) had achieved on other related works.

\subsection{Metrics}\label{sub:metrics}

The metrics used in the experiments were consistent throughout this work. This decision was made to build the ground necessary to compare the results between different experiments. Therefore, two metrics were used in training time and three for the testing step.

\subsubsection*{Training Time}

For the training time, the main metric used was the accuracy metric. Nonetheless, as the model classifies 12 classes, the accuracy reported has two variants: top-1 accuracy and top-5 accuracy (or accuracy@5). 

% Both compute the proportion of the true results (both positive and negative) among the total predictions. However, the first accuracy is interested as the top prediction (using softmax as prediction output) of the model, whereas the second calculates the accuracy among the top 5 predictions. So, if the true label is the second higher prediction, the top-1 accuracy will compute this as an error, on the other hand, the top-5 accuracy will compute as a correct prediction.

% The formula to compute the accuracy is shown in equation \ref{eq:accuracy}. Where $t_p$ is the true positive predictions, $t_n$ the true negatives, and $s$ the total of samples predicted.

% \begin{equation}
% \textnormal{Accuracy} = \frac{\sum{t_p} + \sum{t_n}}{s}
% \label{eq:accuracy}
% \end{equation}

\subsubsection*{Testing Step}

For the testing step, it was created a process that the predictions for both the validation and the test datasets were generated. With these predictions in hand, as well as the true labels of the examples, it was possible to create a confusion matrix for the model. Furthermore, with the confusion matrix at hand, was simple to compute other metrics, such as precision, recall (or sensitivity), and accuracy as well. 

% Both metrics, of recall and precision, can be seen on the equation \ref{eq:prec_recall}. Where $f_p$ is a false positive and $f_n$ is a false negative.

% \begin{equation}
% \textnormal{Precision} = \frac{t_p}{t_p + f_p} \hspace{0.05\textwidth} \textnormal{and} \hspace{0.05\textwidth} \textnormal{Recall} = \frac{t_p}{t_p + f_n}
% \label{eq:prec_recall}
% \end{equation}

Another metric used to evaluate the models was the AUC (Area Under the Curve), along with the ROC (Receiver Operating Characteristic) curve. The ROC curve is a mapping of the sensitivity (probability of detection) versus 1$-$specificity (probability of false alarm), using various thresholds points. Typically, this metric is implemented in systems to analyze how accurately the diagnosis of a patient state is (diseased or healthy) \cite{swets1986indices}. Furthermore, the AUC summarizes the ROC curve and effectively combines the specificity and the sensitivity that describes the validity of the diagnosis \cite{kumar2011receiver}.

Alongside with the ROC curve analysis, is common to calculate the optimal cut-off point. This is used to further separate the test results, so that a diagnosis of diseased or not is provided. When the point is closest to where the sensitivity is equal one and specificity is equal zero, it has achieved the best result possible \cite{hajian2013receiver, unal2017defining}.

% This was made to have a solid material on how the model is performing and what it should improve. Since the recall computes the probability of the detection of a lesion, the precision computes the degree to which a lesion will be classified as their true label (how precise is the recall), and the AUC computes how accurately a diagnosis is being delivered.

\subsection{Best Experiment}

The best results achieved on this work were with the use of the ResNet-152 architecture, trained over an augmented dataset with a mixture of MED-NODE, Edinburgh and Atlas datasets. The augmentation made was of 29 times for each class, leaving the classes unbalanced. 

\begin{figure*}
  \centering
  \includegraphics[width=\textwidth]{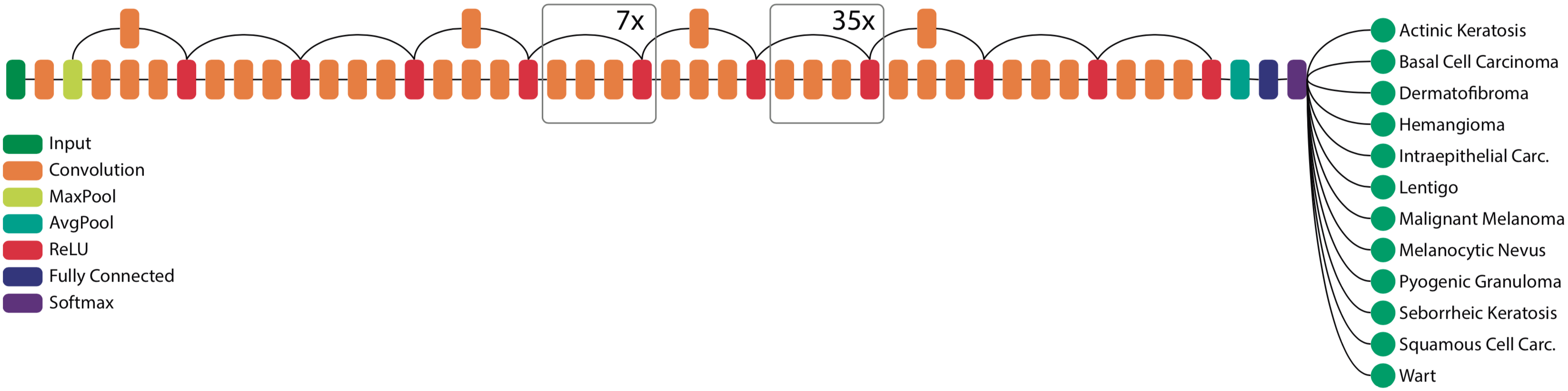}
  \caption{ResNet-152 architecture used.}
  \source{Authors.}
  \label{fig:resnet-152}
\end{figure*}

Furthermore, the ResNet architecture had to be modified to accommodate the needs of the problem at hand. So, the last layer of the architecture was changed from \num{1000} classes to \num{12} classes. Therefore, the final architecture produced followed the same schema seen in Figure \ref{fig:resnet-152}.

Moreover, the technique of transfer learning was applied to generate the best results more rapidly. For that, the hyperparameters of the network had to be tuned and carefully set, for that same purpose.

\subsubsection{Dataset}

The dataset used for the experiment was a derivative of the previous experiment. The original dataset consisted of a mixture of MED-NODE, Edinburgh and Atlas images, moreover, the dataset did not go under augmentation processes and was divided into three separated directories following the division of \num{20}\%, \num{10}\%, and \num{70}\% for testing, evaluation, and training, respectively.

% Furthermore, this dataset was then used for the final experiment. However, by a mistake made in the time of the experiment, only the training and testing directories were used for the experiment. This left \num{533} evaluation images unused, using only \num{3797} images for training and \num{956} for testing. 
Moreover, the training and validation datasets were augmented, using the transformations listed in table \ref{tab:transformations}, by a factor of 29 times. The testing dataset did not suffer any transformations. Finally, the final numbers for the datasets can be seen in table \ref{tab:dataset_used}.

% With this oversight of the evaluation images directory, the training dataset was further divided with a proportion of 80/20 between training and evaluation datasets. 

\begin{table}[htb]
\centering
\caption{Number of images used in the dataset for the final experiment.}
\label{tab:dataset_used}
\begin{tabular}{lrrr}
                          & \multicolumn{3}{c}{\textit{\textbf{Number of images}}} \\ \hline
\textbf{\textit{Lesion Type}} & \textbf{\textit{Train}} & \textbf{\textit{Validation}} & \textbf{\textit{Test}} \\ \hline
Actinic Keratosis         & \num{742}      & \num{186}     & \num{8}             \\
Basal Cell Carcinoma      & \num{30067}    & \num{7517}    & \num{324}           \\
Dermatofibroma            & \num{1067}     & \num{267}     & \num{12}            \\
Hemangioma                & \num{1601}     & \num{400}     & \num{18}            \\
Intraepithelial Carcinoma & \num{1299}     & \num{325}     & \num{14}            \\
Lentigo                   & \num{1137}     & \num{284}     & \num{13}            \\
Malignant Melanoma        & \num{6218}     & \num{1554}    & \num{68}            \\
Melanocytic Nevus (mole)  & \num{17632}    & \num{4408}    & \num{191}           \\
Pyogenic Granuloma        & \num{371}      & \num{93}      & \num{5}             \\
Seborrheic Keratosis      & \num{19256}    & \num{4814}    & \num{208}           \\
Squamous Cell Carcinoma   & \num{1462}     & \num{365}     & \num{16}            \\
Wart                      & \num{7238}     & \num{1810}    & \num{79}            \\ \hline
\textbf{\textit{TOTAL}}   & \num{88090}    & \num{22023}   & \num{956}           %\\ \hline       
\end{tabular}
\end{table}

\subsubsection{Training}

% For the training process, it was used the technique of transfer learning with the approach of fine-tuning the network. 
For the training phase, it was used transfer learning techniques. Thus, it was necessary to gather the ResNet-152 pre-trained weights for the ImageNet dataset\footnote{Available at \url{https://github.com/KaimingHe/deep-residual-networks}. Last accessed on June 26th, 2018.} first, and then modify the network for the purpose of this work.

% Moreover, seeing the work done by \citetextd{SeogHan2018} and the method used to increase the fully connected layer learning rate. 

The learning rate was chosen to be higher than the used in the related works, for two major factors. First of all, one of the early experiments done showed that with a low learning rate it was found a plateau on the very start of the training. Thus, the network did not have the power to learn the features of the skin lesions. Secondly, it was found that increasing the learning rate often aids to reduce underfitting \cite{smith2018hyperparameters}. 
% Therefore, with the past experiments, the number of \num{0.01} was found. However, to the counterpart, the high learning rate, the \textit{stepsize} was decreased to one \textit{epoch}, and a frequent test phase was used to monitor the network learning.

% The batch size was not chosen by experiments, but rather by the hardware limitations. However, the \textit{iter\_size} was chosen to achieve a batch size close to \num{64}, that was a size found to accelerate the training process towards the final results, but not deteriorate the stability of the network \cite{masters2018batchsize}.

Additionally, the final dense layer has a 10 times factor of multiplication for the learning rate, compared to the other layers of the network. However, different from the process of freezing the early layers, used in the same research, this work approximates more to the approach implemented in \cite{Esteva2017}, that fine-tuned all the layers of the network.

This was done with the premise in mind, that although the ImageNet dataset is far diverse and comprehends many different objects, it does not have classes that approximate in characteristics and problems encountered in this dataset of skin lesions. Furthermore, it the weights in the early layers may not be properly trained to extract fine features such as the ones found within the problem that is faced in this work. Therefore, it was needed to fine-tune the learnable parameters since the early layers and learn the final classifier from scratch.

\subsubsection{Hyperparameters}

For this work it was used the Caffe framework \cite{jia2014caffe}, since it allowed and simplified the changes that were needed to do in the layer levels. Furthermore, all the hyperparameters were defined in a separated configuration file called ``Solver'', necessary to define a \textit{.prototxt} file with free parameters used in the training. These hyperparameters can be seen in table \ref{tab:hyperparameters}.

\begin{table}[]
    \centering
    \caption{Hyperparameters used.}
    \label{tab:hyperparameters}
    \begin{tabular}{lr}\hline
        \textbf{Hyperparameter}     &   \multicolumn{1}{r}{\textbf{Value}}      \\ \hline
        \textit{base\_lr}           &   \num{0.01}                              \\
        \textit{weight\_decay}      &   \num{0.00001}                           \\
        \textit{momentum}           &   \num{0.9}                               \\
        \textit{gamma}              &   \num{0.1}                               \\
        \textit{batch\_size}        &   \num{5}                                 \\
        \textit{max\_iter}          &   \num{176180}                            \\
        \textit{test\_iter}         &   \num{22023}                             \\
        \textit{test\_interval}     &   \num{2000}                              \\
        \textit{stepsize}           &   \num{17618}                             \\
        \textit{iter\_size}         &   \num{12}                                \\ \hline
    \end{tabular}
\end{table}

Due to the infrastructure limitations it was only possible to set the \textit{batch\_size} to 5. However, the Caffe framework provides an hyperparameter that serves as a hold on the update of the gradients. The \textit{iter\_size} defines how many iterations the gradients will wait until the update. Altough using this hyperparameter may affect the batch normalization layers used in the architecture, the final results did not show this effect. Furthermore, the maximum iteration parameter was chosen to calculate the number of \textit{epochs}to 10.

\subsubsection{Infrastructure}

All the experiments were conducted under the same environment, that consisted of Antergos 18.3 (Linux kernel 4.16) running BVLC Caffe \cite{jia2014caffe} with support for an NVIDIA GTX 1070 GPU (Cuda 9.1 and cuDNN 7.1).

\section[Results]{Results}\label{ch:results}
% \subsection{Experiments}

Finally, the model used in the testing phase was the product of the iteration number \num{38000}. This training phase took an uninterrupted total time of 35 hours (approximately \num{167} seconds for every 50 iterations).

\subsection{Metric Results}

% For this phase, the model generated in training was submitted to analysis with the testing dataset. Furthermore, the metrics defined in subsection \ref{sub:metrics} were used to analyze the predictions of the model.

With the confusion matrix generated for the predictions in the testing dataset, was found that for all the 11 lesions, with exception of the \textit{Actinic Keratosis}, achieved a accuracy higher than 80\%.
% , using the formula shown in equation \ref{eq:accuracy} (seen on Figure \ref{fig:confusion_matrix} in Appendix \ref{apx:metrics}), 
Thus accounting for a 78\% total accuracy for the model. However, this metric has a bias attached to it, since the distribution of the classes is not even, and therefore can cause misleading in the analysis of this metric. 

% One takeaway from this matrix is the trouble that the model has to predict some class. When in the same row two classes have high color values, it means that there is a high error rate in the row, caused by a confusion between these classes. Thus, it may mean that the two classes have some shared characteristics that cause this phenomenon.

% Furthermore, the classification report was calculated, this gives us the recall, precision and f-1 score for the individual classes as well as the total average. 
% This values can be seen on table \ref{tab:classification_report} in Appendix \ref{apx:metrics}.

\begin{figure}
    \centering
    \begin{subfigure}[t]{0.32\columnwidth}
        \includegraphics[width=\columnwidth]{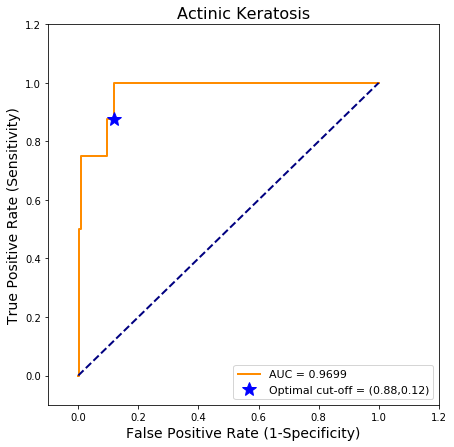}
        \caption{Actinic Keratosis}
        \label{fig:roc_actinic_keratosis}
        ~
    \end{subfigure}
    \begin{subfigure}[t]{0.32\columnwidth}
        \includegraphics[width=\columnwidth]{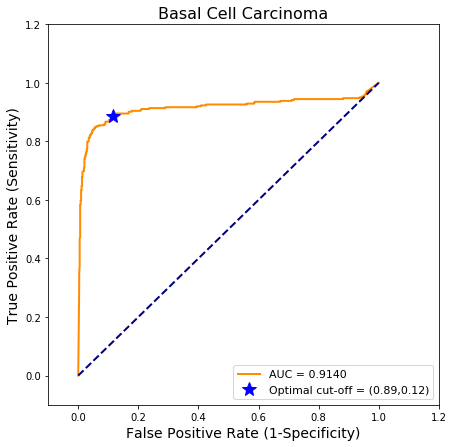}
        \caption{Basal Cell Carcinoma}
        \label{fig:roc_basal_cell_carcinoma}
        ~
    \end{subfigure}
    \begin{subfigure}[t]{0.32\columnwidth}
        \includegraphics[width=\columnwidth]{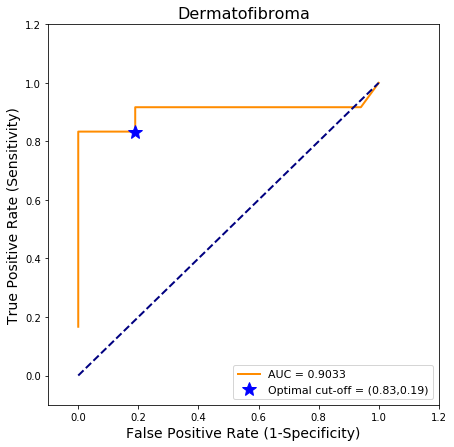}
        \caption{Dermatofibroma}
        \label{fig:roc_dermatofibroma}
        ~
    \end{subfigure}
    % ------
    \begin{subfigure}[t]{0.32\columnwidth}
        \includegraphics[width=\columnwidth]{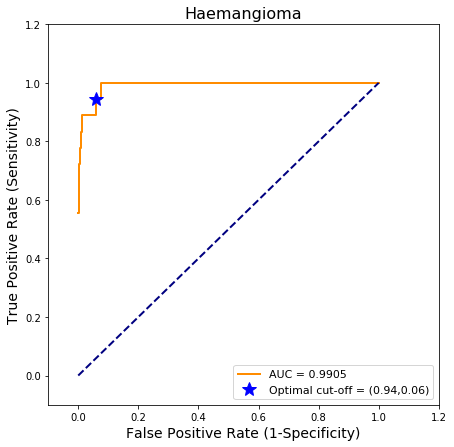}
        \caption{Haemangioma}
        \label{fig:roc_haemangioma}
        ~
    \end{subfigure}
    \begin{subfigure}[t]{0.32\columnwidth}
        \includegraphics[width=\columnwidth]{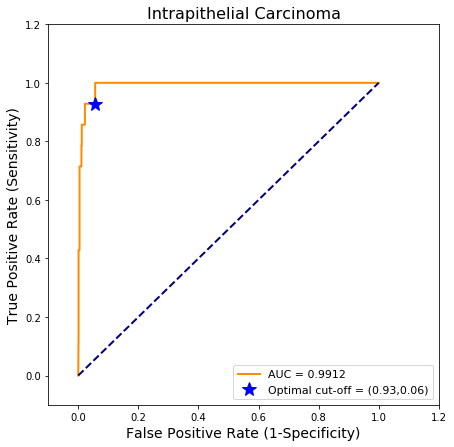}
        \caption{Intrapithelial Carcinoma}
        \label{fig:roc_intrapithelial_carcinoma}
        ~
    \end{subfigure}
    \begin{subfigure}[t]{0.32\columnwidth}
        \includegraphics[width=\columnwidth]{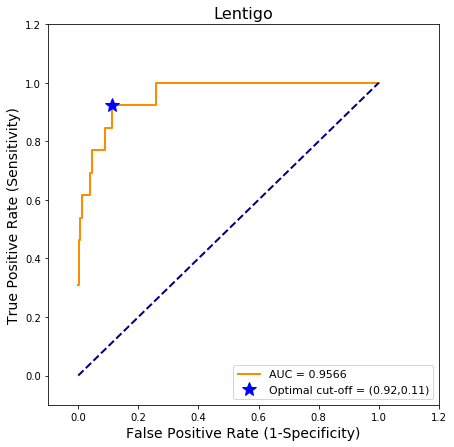}
        \caption{Lentigo}
        \label{fig:roc_lentigo}
        ~
    \end{subfigure}
    % ------
    \begin{subfigure}[t]{0.32\columnwidth}
        \includegraphics[width=\columnwidth]{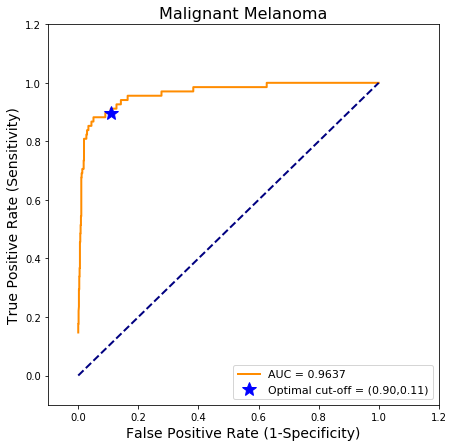}
        \caption{Malignant Melanoma}
        \label{fig:roc_malignant_melanoma}
        ~
    \end{subfigure}
    \begin{subfigure}[t]{0.32\columnwidth}
        \includegraphics[width=\columnwidth]{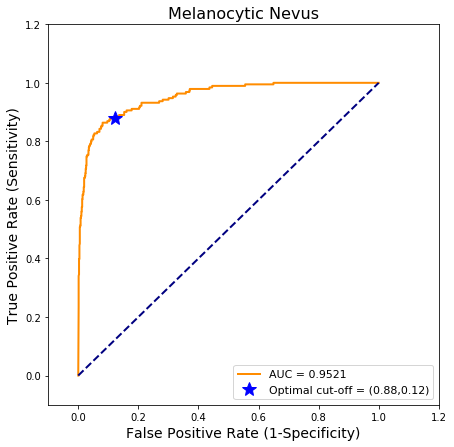}
        \caption{Melanocytic Nevus}
        \label{fig:roc_melanocytic_nevus}
        ~
    \end{subfigure}
    \begin{subfigure}[t]{0.32\columnwidth}
        \includegraphics[width=\columnwidth]{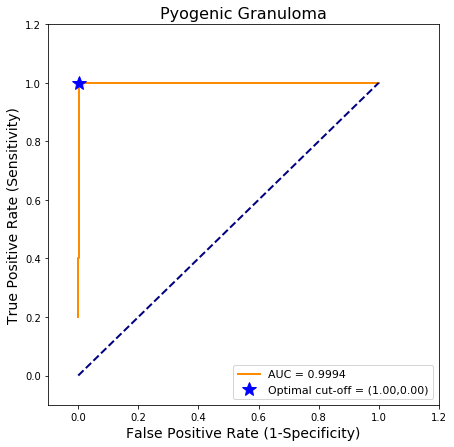}
        \caption{Pyogenic Granuloma}
        \label{fig:roc_pyogenic_granuloma}
        ~
    \end{subfigure}
    % ------
    \begin{subfigure}[t]{0.32\columnwidth}
        \includegraphics[width=\columnwidth]{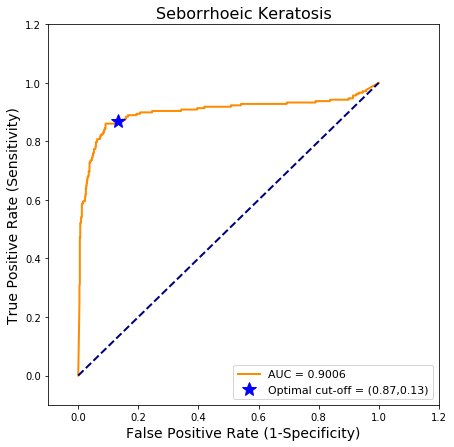}
        \caption{Seborrhoeic Keratosis}
        \label{fig:roc_seborrhoeic_keratosis}
        ~
    \end{subfigure}
    \begin{subfigure}[t]{0.32\columnwidth}
        \includegraphics[width=\columnwidth]{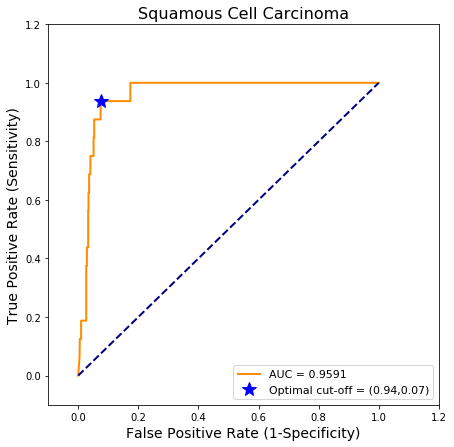}
        \caption{Squamous Cell Carcinoma}
        \label{fig:roc_squamous_cell_carcinoma}
        ~
    \end{subfigure}
    \begin{subfigure}[t]{0.32\columnwidth}
        \includegraphics[width=\columnwidth]{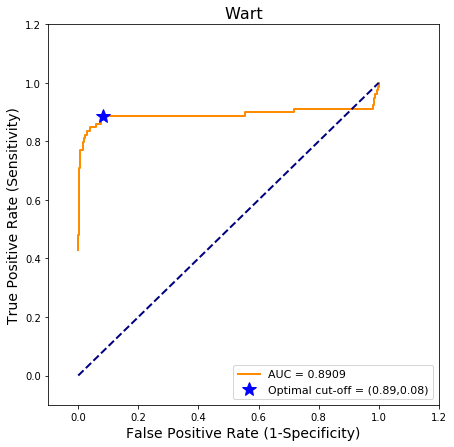}
        \caption{Wart}
        \label{fig:roc_wart}
        ~
    \end{subfigure}
    % ------
    \caption{AUC and Optimal Cut-off for each lesion.}
    \source{Authors.}
    \label{fig:rocs}
\end{figure}

Finally, the AUC and cut-off values for each ROC curve have been calculated (Figure \ref{fig:rocs}). 
% This metric is common among many kinds of research that deals with classification of diagnostics. Moreover, this metric has been used in the researches used as guidelines to quantify the quality of the trained models.
The table \ref{tab:auc_metric} shows a comparison between results in these three works.

% The work done by \citetextd{Esteva2017} was faced with a final binary classification on benign and malignant lesions, thus the metric that was fair to compare was only the AUC metric for Melanoma lesions. However, the work done by \citetextd{SeogHan2018} used a trained model for the exact same lesions, thus it was fair the comparison. Additionally, the results of this work seen on table \ref{tab:auc_metric} can be further examined in graphics on Appendix \ref{apx:metrics}.

\begin{table}[]
\centering
\caption{Comparative between AUC metrics.}
\label{tab:auc_metric}
\begin{tabular}{lccc}\hline
Lesion          & \multicolumn{1}{r}{\citeauthor{Esteva2017}} & \multicolumn{1}{r}{\citeauthor{SeogHan2018}} & \multicolumn{1}{r}{This work} \\ \hline
Actinic Keratosis         & -                                          & 0.83                                        & \textbf{0.96}                          \\
Basal cell carcinoma      & -                                          & 0.90                                        & \textbf{0.91}                          \\
Dermatofibroma            & -                                          & \textbf{0.90}                                        & \textbf{0.90}                          \\
Hemangioma                & -                                          & 0.83                                        & \textbf{0.99}                          \\
Intraepithelial carcinoma & -                                          & 0.83                                        & \textbf{0.99}                          \\
Lentigo                   & -                                          & \textbf{0.95}\footnotemark                           & \textbf{0.95}                          \\
Malignant Melanoma        & \textbf{0.96}                                       & 0.88                                        & \textbf{0.96}                          \\
Melanocytic nevus         & -                                          & 0.94                                        & \textbf{0.95}                          \\
Pyogenic granuloma        & -                                          & 0.97                                        & \textbf{0.99}                          \\
Seborrheic keratosis      & -                                          & 0.89                                        & \textbf{0.90}                          \\
Squamous cell carcinoma   & -                                          & 0.91                                        & \textbf{0.95}                          \\
Wart                      & -                                          & \textbf{0.94}\footnotemark[\value{footnote}]         & 0.89                          \\ \hline
\end{tabular}
\end{table}
\footnotetext{Metric calculated with an Asian dataset, thus may not serve as a comparative in a \textit{stricto sensu}.}

% \subsection{Interpretability}

% The use of artificial intelligence models has been rising in the last few years. Seeing that, the typical use of these models have been similar to black-boxes, where the end-user does not know what it is happening on the insides of the model. This tends to surface some question about whether the machine learning algorithm ``knows'' what is doing, and if ``knows'' better than others that could do the job. 

% This became a serious question when the predictions start to fail. When this happens there is no right answer to what is happening. This could be because of some biases in the training or testing data, or to peculiarities inherent of the algorithm. However, without digging deeper and uncovering the real reasons there is no way to know. For that, the addition of explanations to learning models is needed.

\subsection{Interpretability}

For this work, it was judged to be important to bring an input of the model`s interpretability, since the application is sensible with human life. Moreover, this work want to raise the importance for the use of interpretability techniques for machine learning in medical applications. In addition, an model with an explainability may be more well received by medical practitioners, since it demystify the decisions taken by the model.

Moreover, model interpretability is important at several points. For example, in the training phase if a model is behaving unexpectedly an engineer must know what is happening in order to reverse the situation. For that, if the engineer has in hands the interpretability features of the data, it may found out that the reasons to the behavior might be because of a hindsight bias \cite{fischhoff1975knew} present in the dataset. So, when debugging an application that has a machine learning model, it is important to have the tools to debug properly.

Another fact that is important to discuss is trust. When people are using systems, specially life-critical systems such as medical software, people want to trust that nothing unexpected will happen. And trust for machine learning models can be gained in two ways: through daily use evidence, that can be achieved by getting a high accuracy, for example; another route is to explain how did the model reach the decision, that in this way the end-user will have not only proof that the decision was correct, but how did it know it was a correct decision to make. The second approach can lead to a scenario that the ``black-box'' becomes less so, and less intimidating, thus leading the user to be more inclining to use the prediction to take action \cite{lombrozo2006structure}. According to \citetextd{darpa2016xai}, it is necessary to create a new approach to how the machine learning models present their predictions.

\subsubsection{Preliminary Results}

An analysis of the model predictions was made, as a way to discover which were the examples in the validation dataset, that the model most got right and wrong. This is used as an artifact for interpretability, since it can bring new insights about the heuristics that are causing the model`s decisions. Additionally, it is also commonly used even among simpler applications due to the simplicity of using this technique.

Furthermore, the GradCAM technique \cite{selvaraju2016gradcam} was applied for this work, as a method to have visual feedback to where the network`s activations, previous to the softmax layer, were more predominant. This gives more artifacts to build an explanation for the model decisions. This can also be used to identify problems where the network learns adjacent features that may be present in lesion samples, but does not necessarily hold meaning with the lesion (e.g. nails or hair in a scalp near the lesion), \textit{vide} the ``Husky vs Wolf'' experiment in \cite{ribeiro2016trust}, where background snow was a predominant feature in classifying a Wolf. 

\begin{figure}
    \centering
    \begin{subfigure}[t]{\columnwidth}
        \includegraphics[width=\columnwidth]{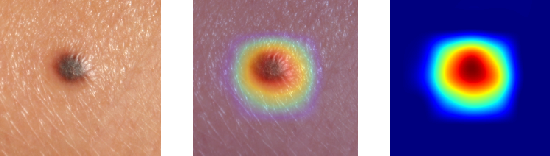}
        \caption{Predicted as Haemangioma with 100\% confidence.}
        \source{Edinburgh dataset.}
        \label{fig:melanoma_haem_wrong}
        ~
    \end{subfigure}
    
    \begin{subfigure}[t]{\columnwidth}
        \includegraphics[width=\columnwidth]{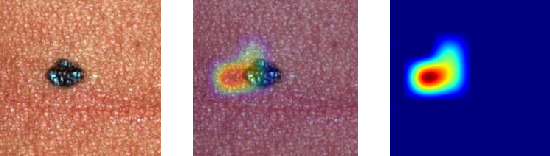}
        \caption{Predicted as Basal Cell Carcinoma with 100\% confidence.}
        \source{Edinburgh dataset.}
        \label{fig:melanoma_bcc_wrong}
        ~
    \end{subfigure}
    
    \begin{subfigure}[t]{\columnwidth}
        \includegraphics[width=\columnwidth]{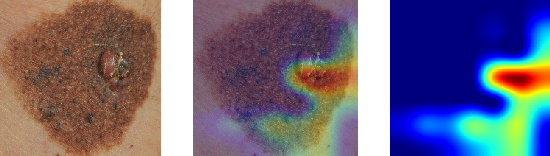}
        \caption{Predicted as Melanocytic Nevus with 97\% confidence.}
        \source{MED-NODE dataset.}
        \label{fig:melanoma_nevus_wrong}
    \end{subfigure}
    \caption{GradCAM applied to the most wrong predictions for Malignant Melanoma. Columns from left to right: original image; original image fused with heat-map; heat-map produced by GradCAM.}
    \label{fig:melanoma_wrong}
\end{figure}

For the most wrong predictions, it was found that the causes may fall under 2 factors: the lesion analyzed indeed caused confusion between the lesions (Figure \ref{fig:melanoma_haem_wrong}); the model did not generalize well and was struggling to extract predominant features in some of the images, thus giving more importance to areas that were not relevant, from a practical perspective (Figure \ref{fig:melanoma_bcc_wrong} and \ref{fig:melanoma_nevus_wrong}). 
% These pictures can be seen on Appendix \ref{apx:interpretability}.

\begin{figure}
    \centering
    \begin{subfigure}[t]{\columnwidth}
        \includegraphics[width=\columnwidth]{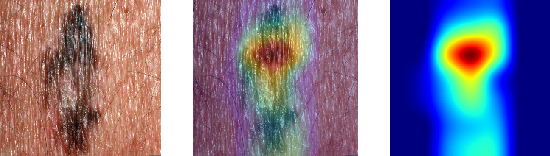}
        \caption{Melanoma with 100\% confidence.}
        \source{Edinburgh dataset.}
        \label{fig:melanoma_correct_1}
        ~
    \end{subfigure}
    
    \begin{subfigure}[t]{\columnwidth}
        \includegraphics[width=\columnwidth]{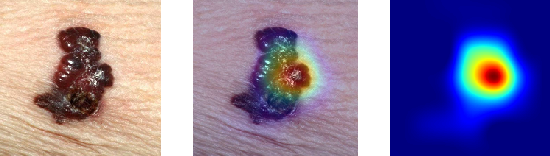}
        \caption{Melanoma with 100\% confidence.}
        \source{Edinburgh dataset.}
        \label{fig:melanoma_correct_2}
        ~
    \end{subfigure}
    
    \begin{subfigure}[t]{\columnwidth}
        \includegraphics[width=\columnwidth]{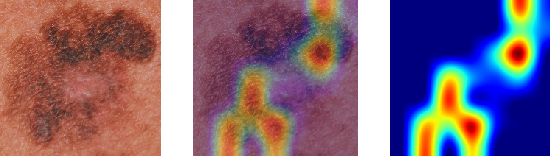}
        \caption{Melanoma with 100\% confidence.}
        \source{MED-NODE dataset.}
        \label{fig:melanoma_correct_3}
    \end{subfigure}
    \caption{GradCAM applied to the most correct predictions for Malignant Melanoma. Columns from left to right: original image; original image fused with heat-map; heat-map produced by GradCAM.}
    \label{fig:melanoma_correct}
\end{figure}

The most correct lesions brought more information on what the model was already good at, and how this translate in a perspective of image features. For example, in Figures \ref{fig:melanoma_correct_1} and \ref{fig:melanoma_correct_2}, the model found out that the regions that mostly identify the lesions as Malignant Melanomas are indeed the ones that would bring more relevance to the doctor`s decision making process. However there are still some examples, such as Figure \ref{fig:melanoma_correct_3}, that are more emblematic and need a expert`s eye to shed a light on it. Moreover, we can speculate that the model took advantage of the geometric and color asymmetry in the lesion to make an accurate decision.

% The undecided lesions brought more information to what the model is struggling with, that is necessary to further dedicate more time to the analysis of what it might be causing the undecided cases such as Figures \ref{fig:nevus_melanoma_und} and \ref{fig:bcc_nevus_und}.

% Furthermore, the GradCAM method was applied for this work, however, for that, it was necessary to implement the code to utilize it in python alongside with Caffe, since the official repository\footnote{Available at \url{https://github.com/ramprs/grad-cam/}. Last accessed in 28/06/2018.} utilizes LUA language for its solution. 
% This implementation can be seen on Appendix subsection \ref{apxsb:gradcam}.

Furthermore, it was found that the model generalizes well for the examples that it was shown, correctly activating the regions that contained the lesion, even in images that the pose made challenging the localization of the lesion. 
% Moreover, the Figure \ref{fig:gradcam_lesion} shows an example of such pose, that the nose in an acute angle distances the lesion from focus, in spite of this, the model is able to detect not only the class but the localization of the lesion.

% \begin{figure}
%   \centering
%   \includegraphics[width=\columnwidth]{images/grad_cam.png}
%   \caption{\textit{Basal Cell Carcinoma} lesion with 100\% confidence.}
%   \source{Author.}
%   \label{fig:gradcam_lesion}
% \end{figure}

% This technique makes especially easy to detect whether the model has generalized well over the problem or if it is detecting biases in the dataset, such as some kind of skin texture. 
Nonetheless, this kind of interpretation is important not only in a developer vision but also for the possible doctors that were to receive a simple prediction and to take action on another human life based on it. With these other tools, the doctor has much more to support the next decision that is necessary to take on the patient. Therefore, making the models as interpretable as accurate can transform a tool in a good counselor.

\section{Discussions}

In this paper, we discussed the importance of automatic classification method to support skin lesions diagnosis. Furthermore, we listed a group of researches and their achieved results for the same problem. However, it is still a problem with several difficulties, even more when we study clinical images, that may present an immense diversity due to variables such as cameras and environments.

Seeing this, this work presented a model capable of classifying 12 skin lesions, that reached results comparable with state-of-the-art. Additionally, it was presented studies on the model decision taking process with interpretability techniques. However, regardless of the excellent results encountered in this work, it is necessary to further test the model with more data, with more diversity (different ethnics and ages), and then investigate the results for improvements.

% Another fact that it is worth mentioning is that, although the number of maximum iterations used was a 10 \textit{epochs} iteration size, the training was not concluded after the full completion of these iterations. The training was early stopped, since, around the iteration number \num{30000} the loss function, both in training and validation, did not alter significantly. Therefore, it was judged that the horizontal part of the validation loss was achieved, thus a good convergence of the network \cite{smith2018hyperparameters}. 

% Regardless of the prominent results encountered in this work, it is necessary to further test the model with more data, and with more diversity. Only with that, the final results will be trustworthy.

\bibliographystyle{IEEEtranN}
\bibliography{references}

\end{document}